\documentclass{article} 

 \pdfoutput=1
 
\usepackage{graphicx} 
\usepackage{subfigure} 

\usepackage{natbib}

\usepackage{algorithm}
\usepackage{algorithmic}

\usepackage{amsmath}

\usepackage{amssymb}

 \usepackage{url}

\title{Stochastic gradient variational Bayes for gamma approximating distributions}

\author{David A. Knowles \\ Stanford University \\ dak33@stanford.edu}

\begin{document} 

\maketitle

\begin{abstract}
While stochastic variational inference is relatively well known for scaling inference in Bayesian probabilistic models, related methods also offer ways to circumnavigate the approximation of analytically intractable expectations. The key challenge in either setting is controlling the variance of gradient estimates: recent work has shown that for continuous latent variables, particularly multivariate Gaussians, this can be achieved by using the gradient of the log posterior. In this paper we apply the same idea to gamma distributed latent variables given gamma variational distributions, enabling straightforward ``black box'' variational inference in models where sparsity and non-negativity are appropriate. We demonstrate the method on a recently proposed gamma process model for network data, as well as a novel sparse factor analysis. We outperform generic sampling algorithms and the approach of using Gaussian variational distributions on transformed variables. 
\end{abstract} 

\section{Introduction}

Bayesian probabilistic models offer a clean, interpretable methodology for applied statistical analysis. However, inference remains a challenge both in terms of ease of implementation and scalability. Ease of implementation is important so that practitioners can construct models tailored specifically to their application, rather than being forced to choose from a small set of pre-existing models. While various software packages, such as Infer.NET \citep{minkainfer}, WinBUGS \citep{lunn2000winbugs}, Church \citep{goodman2012church} and more recently, STAN \citep{stan}, have been designed explicitly to address this problem, they do not currently scale to large real world datasets. 

Stochastic variational inference (SVI) methods \citep{hoffman2010online,hoffman2013stochastic}, follow the traditional variational Bayes approach of converting an intractable integration into a optimization problem. However, where VB would usually proceed by the well known coordinate ascent updates on the variational lower bound \citep{jordan1999introduction}, SVI utilizes the key idea of stochastic gradient descent: that it is enough to follow noisy, but unbiased, estimates of the gradient \citep{robbinsmonro}. If these noisy gradients can be computed much more cheaply than full gradients, e.g. by subsampling the data, then more rapid convergence is typically possible and the volume of data which can be handled is greatly increased. 

The observation that noisy, unbiased gradients can be used in variational inference suggests another idea: instead of analytically calculating the required expectations and gradient of the lower bound can we just use Monte Carlo? The challenge in applying this idea is to keep the variance of the Monte Carlo gradient estimates low without requiring a computationally infeasible number of samples. Various ``tricks'' have been proposed to achieve this, including control variates \citep{paisleyvb}, stochastic linear regression \citep{salimans2013fixed} and using the factor graph structure of the model \citep{ranganath2013black}. We focus on a recently proposed solution for continuous latent variables proposed independently by \citet{salimans2013fixed} and \citet{kingma2013auto} which utilizes just the gradient of the log posterior, which we refer to as stochastic gradient variational Bayes (SGVB). While both papers demonstrated the effectiveness of this approach for multivariate Gaussian variables, whether it is equally useful for latent variables with very different distributions remains an open question. In this paper we investigate using gamma approximating distributions. Compared to the Gaussian case, gamma r.v.s represent a natural step in the direction of more structured models: despite being continuous they can encode sparsity using suitably small shape parameters, while also enforcing non-negativity, which is appropriate in many settings. In addition gamma r.v.s also underly many of the most commonly used Bayesian nonparametric priors such as the Dirichlet process. 

The variational autoencoder \citet{kingma2013auto} uses a variational inference methodology where the approximate posterior is a function, known as the recognition model, of the observed data. This allows extremely scalable training using stochastic gradient descent analogously to a standard autoencoder.  While having certain advantages, this approach can be sensitive to the choice of recognition model, has only been demonstrated for Gaussian latent variables, does not straightforwardly handle missingness in the observations and only performs MLE over the model parameters. 

Related methodology has very recently been incorporated into Stan~\citep{kucukelbir2015automatic}. Their approach is to always use a fully factorized Gaussian variational posterior but to reparameterize such that the space of the r.v.s is always the reals. For r.v.s constrained to be positive for example, this corresponds to using a log-normal variational posterior. Our experiments here suggest that explicitly using gamma variational posteriors, at least when the \emph{priors} are gamma, is preferable. 


%

In Section~\ref{secmethods} we review stochastic gradient variational Bayes (SGVB), show how to leverage the gradient wrt to the log joint and present the necessary derivations for gamma r.v.s. We present two models in Section \ref{sec:models} which we use as test cases. The first is the infinite edge partition model \citep{zhou2015infinite} for network data, the second a novel gamma process factor analysis model (GPFA) for arbitrary continuous data. In Section \ref{sec:results} we present promising results for both models on synthetic and real world data and conclude in Section \ref{sec:conclusion} with some potential future directions. 

\section{Methods} \label{secmethods}

In this section we review variational inference, show how the required gradients can be approximated using Monte Carlo and then turn to the particular case of gamma r.v.s. 

\subsection{Variational inference} 

Let the normalized distribution of interest be $p(\mathbf{x}) = f(\mathbf{x})/Z$. Typically $p$ is the posterior, $f$ is the joint and $Z$ is the marginal likelihood (evidence). We use Jensen's inequality to lower bound
\begin{align}
\log{Z} &= \log \int_\mathbf{x} f(\mathbf{x}) d\mathbf{x} = \log \int_\mathbf{x} q(\mathbf{x}) \frac{f(\mathbf{x})}{q(\mathbf{x})} d\mathbf{x} 
\geq \int_\mathbf{x} q(\mathbf{x}) \log \frac{f(\mathbf{x})}{q(\mathbf{x})} d\mathbf{x} =: \mathcal{F}[q] .
\end{align}
where $q$ represents the variational posterior. We can ask what error we are making between the true $Z$ and the Evidence Lower BOund (ELBO) $\mathcal{F}[q]$:
\begin{align}
\log{Z}-\mathcal{F}[q] = \int_\mathbf{x} q(\mathbf{x}) \log \frac{q(\mathbf{x})}{p(\mathbf{x})} d\mathbf{x} =: KL(q||p) = -H[q(\mathbf{x})] - \int q(\mathbf{x}) \log p(\mathbf{x}) d\mathbf{x},  \label{eq:kl}
\end{align}
where $KL(q||p) $ is the KL divergence and $H[q(\mathbf{x})]=-\int q(\mathbf{x}) \log q(\mathbf{x}) d\mathbf{x}$ is the entropy. In general we can evaluate $\mathcal{F}[q]$ but not the KL itself, since this would require knowing $Z$. By maximising the lower bound $\mathcal{F}[q]$ we will minimize the KL divergence. 
The KL divergence is strictly positive for $q \neq p$ and equal to $0$ only for $q=p$. As a result finding the general $q$ which minimizes the KL divergence is no easier than the original inference task, which we assume is intractable. The usual strategy therefore is to place simplifying constraints on $q$, the most popular, due to its simplicity, being the mean field approximation. We will take the approach of choosing $q$ to have a specific parametric form, indexed by $\theta$. Typically $q_\theta$ will be in the exponential family: in this paper in particular, $q_\theta$ will be a product of gamma distributions. 

\subsection{SGVB for continuous latent variables}

To fit $q_\theta$ we will maximize $\mathcal{F}[q_\theta]$ wrt to $\theta$, which requires estimating the gradient
\begin{align} \nabla_\theta \mathcal{F}[q_\theta] = \nabla_\theta \mathbb{E}_{q_\theta} [ \log f(\mathbf{x}) - \log q_\theta(\mathbf{x}) ] \end{align}
This form is not easily amenable to Monte Carlo estimation because of the dependence of $q_\theta$ on $\theta$. One approach is to use the identity $\nabla_\theta \mathbb{E}_{q_\theta}[ L(\mathbf{x}) ] = \mathbb{E}_{q_\theta}[  L(\mathbf{x}) \nabla_\theta \log q_\theta(\mathbf{x}) ]$ where $L(\mathbf{x})=\log f(\mathbf{x}) - \log q_\theta(\mathbf{x})$, but this typically has high variance. Instead, assume we can find a random variable $\mathbf{z} \sim \pi(\mathbf{z})$ such that $\mathbf{x}=\psi(\mathbf{z},\theta)$ has the same distribution as $\mathbf{x} \sim q_\theta$, then
\begin{align}
\nabla_\theta \mathbb{E}_{q_\theta}[ L(\mathbf{x}) ] = \mathbb{E}_{\pi(\mathbf{z})} [ \nabla_\theta \psi(\mathbf{z},\theta) \nabla_\mathbf{x} L(f(\mathbf{z},\theta)) ].
\end{align}
Since $\pi(\mathbf{z})$ has no dependence on $\theta$ the RHS expression is straightforward to approximate by Monte Carlo,
\begin{align*} 
\nabla_\theta \mathbb{E}_{q_\theta}[ L(\mathbf{x}) ] &\approx \frac{1}{S} \sum_{s=1}^S \nabla_\theta \psi(\mathbf{z}^{(s)},\theta) \nabla_\mathbf{x} L(\mathbf{x}^{(s)}), \quad \text{ where } \quad \mathbf{z}^{(s)} \sim \pi, \mathbf{x}^{(s)} = \psi(\mathbf{z}^{(s)},\theta). 
\end{align*}
In fact, this estimator generally has low enough variance that we can simply use $S=1$. 

For Gaussian random variables $\mathbf{x}$ an obvious choice is $\pi(\mathbf{z})=N(0,I)$ and $\psi(\mathbf{z}, \{ \mathbf{m}, \mathbf{V} \})=\mathbf{m}+\mathbf{V}^\frac12 \mathbf{z}$ where $\{ \mathbf{m}, \mathbf{V} \}$ are the mean and (co)variance respectively. For gamma random variables no such simple transformation exists, so we resort to the generic CDF transform instead. For any random variable $x$ with CDF $F_\theta(x)$ we can sample $x$ as
\begin{align} z \sim U[0,1], x=F^{-1}_\theta(z) =: \psi(z,\theta) \end{align} 
where $U[0,1]$ is the uniform distribution on $[0,1]$. We can differentiate $\psi(z,\theta)$ with respect to $\theta$ as
\begin{align} \label{eq:difficdf}
 \nabla_\theta \psi(z,\theta) = - \frac{ \nabla_\theta F_\theta(x) }{ f_\theta(x) } 
\end{align}
where $ f_\theta(x) $ is the pdf of $x$. 

\subsection{Gamma variational distributions} 

For a gamma latent variable with shape $a$ and rate $b$
\begin{align} F_{a,b}(x) = \int_0^x \frac{b^a}{\Gamma{(a)}} t^{a-1} e^{-bt} dt. \end{align}
It is straightforward to differentiate this expression wrt $b$ and use Equation \ref{eq:difficdf} to obtain $\nabla_b \psi(z,a,b)$. However the result is easier to obtain by noting that $ x = F^{-1}_{a,b}(z)  = F^{-1}_{a,1}(z) / b $ and so
\begin{align}
 \nabla_b \psi(z,a,b) = \nabla_b F^{-1}_{a,b}(z) = \nabla_b F^{-1}_{a,1}(z) / b = - F^{-1}_{a,1}(z) / b^2 = - x / b 
\end{align}
Unfortunately the gradient wrt to the shape $a$ has no analytical form in terms of commonly available special functions. Depending on the order of magnitude of $a$ different approaches can be used to accurately and efficiently approximate $\nabla_a \psi(z,a,b)$. For moderate values of $a$ we use a finite difference approximation
\begin{align} \nabla_a F^{-1}_{a,b}(z) \approx \frac{ F^{-1}_{a+\epsilon,b}(z) - F^{-1}_{a,b}(z) }{ \epsilon } ,\end{align}
where $\epsilon$ is a small positive constant. We use of the high numerical precision of the \texttt{gaminv} Matlab function (the Boost C++ library also implements such a function), which calculates $F^{-1}$ using an iterative solver. 

For small values of $a \ll 1 $ \texttt{gaminv} often fails to converge. This regime is important because it corresponds to the gamma distribution's ability to model sparsity. Fortunately, in this regime the asymptotic approximation $ F_{a,1}(x) \approx \frac{x^a}{a\Gamma(a)} $ becomes increasingly accurate, so that
\begin{align} \label{eqsmalla}
 F^{-1}_{a,b}(z) \approx \left( z a \Gamma(a) \right)^\frac1{a} / b 
 \end{align}
For $a<1$ and $(1-0.94z)\log(a)<-0.42$ we use Equation \ref{eqsmalla} to efficiently obtain both $x$ and $\nabla_a \psi(z,a,b)$ without expensive calls to \texttt{gaminv}, whilst keeping the absolute relative error below $10^{-4}$. Finally for large $a \gg 1$ (we use $a>1000$) the gamma distribution is well approximated by a Gaussian with matched mean and variance, i.e. $\psi(z',a,b) \approx (a+\sqrt{a}z')/b$ and $\nabla_a \psi(z',a,b) \approx (1+z'/\sqrt{a})/b$ where $z'\sim N(0,1)$. 

\subsection{Optimization}

The shape and rate parameters for the gamma distribution are of course required to be positive. To cope with this we use the reparameterisation $r(\theta)=\log(1+\exp{(\theta}))$ for both the shape and rate to avoid performing constrained optimisation. Pseudocode is shown in Algorithm \ref{algo} for the basic algorithm. We also experimented with incorporating momentum, and using AdaGrad \citep{duchi2011adaptive}, RMSprop \citep{rmsprop} or AdaDelta \citep{zeiler2012adadelta} to set the learning rate. Momentum involves maintaining an additional velocity vector $v$ which is updated as $v \leftarrow \lambda g + (1-\lambda) v$ where $g$ is the gradient and $\lambda \in [0,1]$ is the momentum parameter. $v$ is then used in the place of $g$ when updating the parameters. Our implementation of AdaGrad uses a step-size $\gamma^{(t)} = 0.1/\left(10^{-6} + \sqrt{ \sum_{j=1}^t g^2_t} \right)$ where $g_t$ is the gradient at step $t$. RMSprop is similar in spirit to AdaGrad: we maintain a running average $m \leftarrow 0.1 g^2 + 0.9 m$, and use a step-size $\gamma^{(t)} = 0.01/\left(1\times 10^{-6} + \sqrt{m} \right)$. AdaDelta is a heuristic which tries to maintain progress in later stages of the optimization by keeping the same running average of squared gradients as RMSprop, $m_g \leftarrow \rho g^2 + (1-\rho) m_g$ (with $\rho \in [0,1]$), as well as $m_\theta \leftarrow \rho (\Delta \theta)^2 + (1-\rho) m_\theta $ (where $\Delta \theta$ is the update in parameter space), and using a step-size $\gamma=\sqrt( m_\theta + \epsilon ) / \sqrt( m_g + \epsilon ) $, where we use $\epsilon=10^{-4}$. 

\begin{algorithm}[tb]
   \caption{Gamma stochastic gradient variational Bayes}
   \label{algo}
\begin{algorithmic}
   \STATE Initialize $t=0$, $\mathbf{a},\mathbf{b}$.
   \STATE $\boldsymbol{\alpha} = r^{-1}(\mathbf{a}), \boldsymbol{\beta} = r^{-1}(\mathbf{b})$
   \REPEAT
   \STATE Sample $z_d\sim U[0,1]$
   \STATE Set $x_d = F^{-1}_{a_d,b_d}(z_d)$ according to Section 2.3
   \STATE Set $\mathbf{g}=\nabla_\mathbf{x} [ \log f(\mathbf{x}) - \log q_{a,b}(\mathbf{x})]$
   \STATE Set $g_d^\alpha = g_d \nabla_a F^{-1}_{a_d,b_d}(z_d) / (1+e^{-\alpha_d}) $
   \STATE Set $g_d^\beta = g_d \nabla_b F^{-1}_{a_d,b_d}(z_d) / (1+e^{-\beta_d}) $
   \STATE Compute step size $\gamma^{(t)}$ (e.g. using AdaDelta on $[ g^\alpha, g^\beta]$))
   \STATE $\alpha_d \leftarrow \alpha_d + \gamma^{(t)} g^\alpha$
   \STATE $\beta_d \leftarrow \beta_d + \gamma^{(t)} g^\beta$
   \STATE $\mathbf{a}=r(\boldsymbol{\alpha}), \mathbf{b}=r(\boldsymbol{\beta})$
   \STATE $t := t+1 $
   \UNTIL{convergence}
\end{algorithmic}
\end{algorithm}

\section{Models} \label{sec:models}

In this section we briefly outline the models we will use to assess the algorithm. We choose models which only involve gamma latent variables, but emphasize that models involving both Gaussian and gamma latent variables would also be straightforward to implement. 

\subsection{Infinite edge partition model} 

There has been considerable recent interest in probabilistic modelling of network data, typically represented as an undirected graph \citep{Kemp06learningsystems,BluTeh2013a},. In a social network nodes will represent individuals and edges friendships, or in a protein interaction network nodes represent proteins and edges physical interaction. Let $Y \in \{0,1\}^{N \times N}$ represent the binary adjacency matrix of the graph: $y_{ij}=y_{ji}$ indicates whether there is a link between nodes $i$ and $j$. Many models have been proposed to uncover the latent structure in such data, but we will focus on the recent infinite edge partition model \citep[EPM,][]{zhou2015infinite}, which specifies
\begin{align} P( y_{ij}=1 | W ) = 1 - \exp{ \left( -\sum_k r_k w_{ik} w_{jk} \right) } \end{align} 
where $ W $ is a $N \times K$ matrix of positive reals and $r$ is a $K$-vector of reals. 
This link function can be interpreted as summing over latent variables ${ s_{ijk} \sim \text{Poisson}(r_k w_{ik} w_{jk}) } $  and taking $ y_{ij}= \mathcal{I}[ 0 < \sum_{k=1}^K s_{ijk} ] $. This link function has two advantages over logistic link functions: \emph{i)} it is appropriate for sparse graphs since $P(y_{ij}=1 | W )$ is small when $W \approx 0$, \emph{ii)} the corresponding likelihood can be evaluated with computational cost linear in the number of observed present and missing edges (as noted by \citet{morup2011infinite}), which is typically orders of magnitude smaller than $N^2$. To see this note that the likelihood is
\begin{align}
 & \sum_{i>j} M_{ij} [ Y_{ij} \log{ ( 1-e^{ -p_{ij} } ) } +  (1-Y_{ij})(-p_{ij}) ] \notag \\ =  & \sum_{i>j}  M_{ij} Y_{ij} [ \log{ ( 1-e^{ -p_{ij} } )} + p_{ij} ] + \sum_{i>j}  (1-M_{ij}) p_{ij} - \sum_{i>j}  p_{ij} 
 \end{align}
 where $M_{ij}=1$ iff edge $ij$ is not missing and $0$ o.w., and $p_{ij}=\sum_k w_{ik} w_{jk} $. The first sum only involves non-missing existing edges, the second sum only missing edges, and the third sum can be calculated efficiently in $O(NK)$ as
 \begin{align} \sum_{i>j}  p_{ij} = \sum_{i>j}  \sum_k w_{ik} w_{jk} = \frac12 \sum_k [w_{\cdot k}^2 - \sum_i w_{ik}^2] 
 \end{align}
 where $ w_{\cdot k} = \sum_i w_{ik} $. 
 
 To complete the prior specification we use 
\begin{align}
 W_{ik}|a_i,c_i \sim G(a_i, c_i), \quad a_i \sim G(0.01, 0.01),  \quad c_i \sim G(1,1), \notag \\ 
  r_k|\gamma_0,c_0 \sim G( \gamma_0/K, c_0),  \quad  \gamma_0 \sim G(1,1),  \quad c_0 \sim G(1,1) 
  \end{align} 
 where the distribution on $r_k$ is a finite $K$ approximation to the gamma process. 

\subsection{Gamma process factor analysis} \label{secGpfaModel}

Factor analysis models are appealing for finding latent structure in high dimensional data. Observed data samples $\mathbf{y}_n \in \mathbb{R}^D,  n=1 \dots N$ are modeled as
\begin{align} \mathbf{y}_n | \mathbf{x}_n \sim N( \mathbf{W} \mathbf{x}_n, \sigma^2 \mathbf{I} ) \end{align}
where typically $\mathbf{x}_n \sim N(0,\mathbf{I})$. Many approaches exist to fitting such models. From a Bayesian viewpoint, if $\mathbf{W}$ is given a Gaussian prior then Gibbs sampling $x|W,-$ and $W|x,-$ is straightforward and conjugate, although even in this simple setting the strong posterior dependencies between $x$ and $W$ can be problematic for convergence and mixing. We consider placing a Gamma prior on the elements of $W$, thereby enforcing positivity. Such ``semi"-nonnegative matrix factorization (NMF) is closely connected to $k$-means clustering \citep{ding2010convex}, encouraging interpretable solutions while still allowing arbitrary real valued data, unlike classical NMF which requires positive data. Since our SGVB algorithm does not require conjugacy, we can integrate out $\mathbf{x}_n$ to give 
\begin{align} \mathbf{y}_n \sim N( \mathbf{0}, \mathbf{W} \mathbf{W}^T +  \sigma^2 \mathbf{I} ) \end{align}
The log likelihood is then 
\begin{align} L= \frac{N}2 \log{ | \mathbf{K} | } - \frac12 \sum_i \mathbf{y}_i^T \mathbf{K}^{-1}  \mathbf{y}_i = \frac{N}2 \log{ | \mathbf{K} | } - \frac12 \text{tr}( YY^T \mathbf{K}^{-1})\end{align}
where $ \mathbf{K} = \mathbf{W} \mathbf{W}^T +  \sigma^2 \mathbf{I} $ and $ Y=[ \mathbf{y}_1, ... , \mathbf{y}_N ]$. Differentiating w.r.t. $\mathbf{W}$ we have
\begin{align} \frac{\partial L}{\partial \mathbf{W}} = N\mathbf{K}^{-1} \mathbf{W} + \mathbf{K}^{-1}YY^T\mathbf{K}^{-1}\mathbf{W} \end{align} 
After precomputing $YY^T$ the per iterations operations are $O(D^3)$, with no dependence on $N$. Similarly to the EPM we use a hierarchical gamma process prior construction for $W$: 
\begin{align}
W_{dk}|r_k,\gamma \sim G( \gamma r_k, \gamma), &\quad \gamma \sim G(1,1), \notag \\
 r_k|\gamma_0,c_0 \sim G( \gamma_0/K, c_0 ),  &\quad  \gamma_0 \sim G(1,1),  \quad c_0 \sim G(1,1).
 \end{align}

\section{Results} \label{sec:results}

We present results on both synthetic and real world data for the two models described in Section \ref{sec:models}, with inference performed using our gamma SGVB algorithm. 

\subsection{Infinite edge partition model} 

\begin{figure}[htbp]
\begin{center}
\includegraphics[width=.8\textwidth]{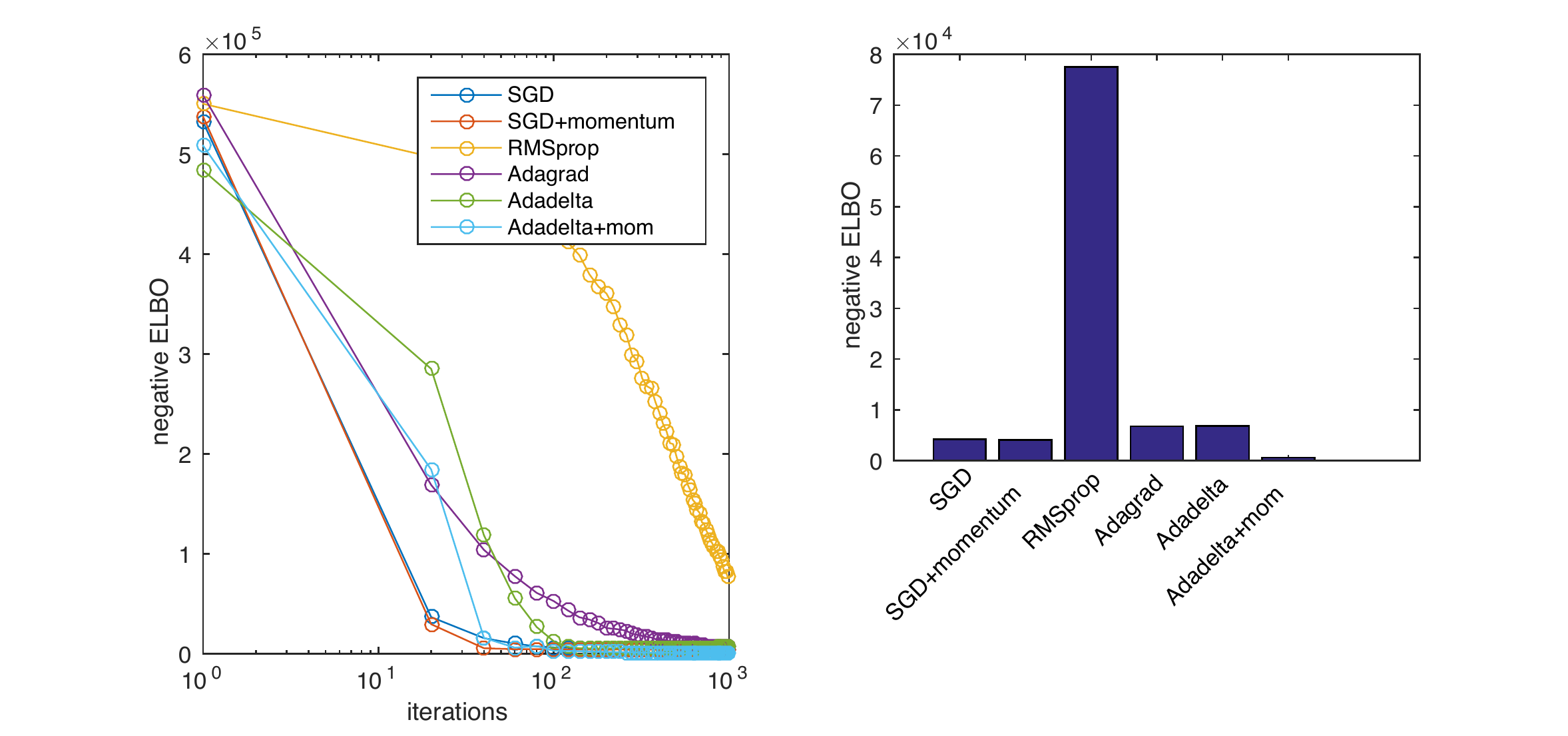}
\caption{Performance of various learning rate adaption methods, including using momentum, for GammaSGVB. \textbf{Left}: Negative ELBO (lower is better) with \# iterations. \textbf{Right:} Final ELBO after 1000 iterations. }
\label{networkCompareStep}
\end{center}
\end{figure}

We initially investigated what choices of step size adaptation and momentum were most compatible with the gamma SGVB algorithm (Figure \ref{networkCompareStep}), at least in the context of the EPM. All methods performed comparably apart from RMSprop which performed poorly in this setting. Adadelta with momentum, despite not achieving the fastest initial improvement, obtained the best ELBO after 1000 iterations, likely because Adadelta continues to make progress after Adagrad and standard SGD have stopped, and because momentum helps smooth over the stochasticity of the gradient estimates. 

\begin{figure}[htbp]
\centering
\begin{minipage}{0.45\textwidth}
\begin{center}
\includegraphics[width=\textwidth]{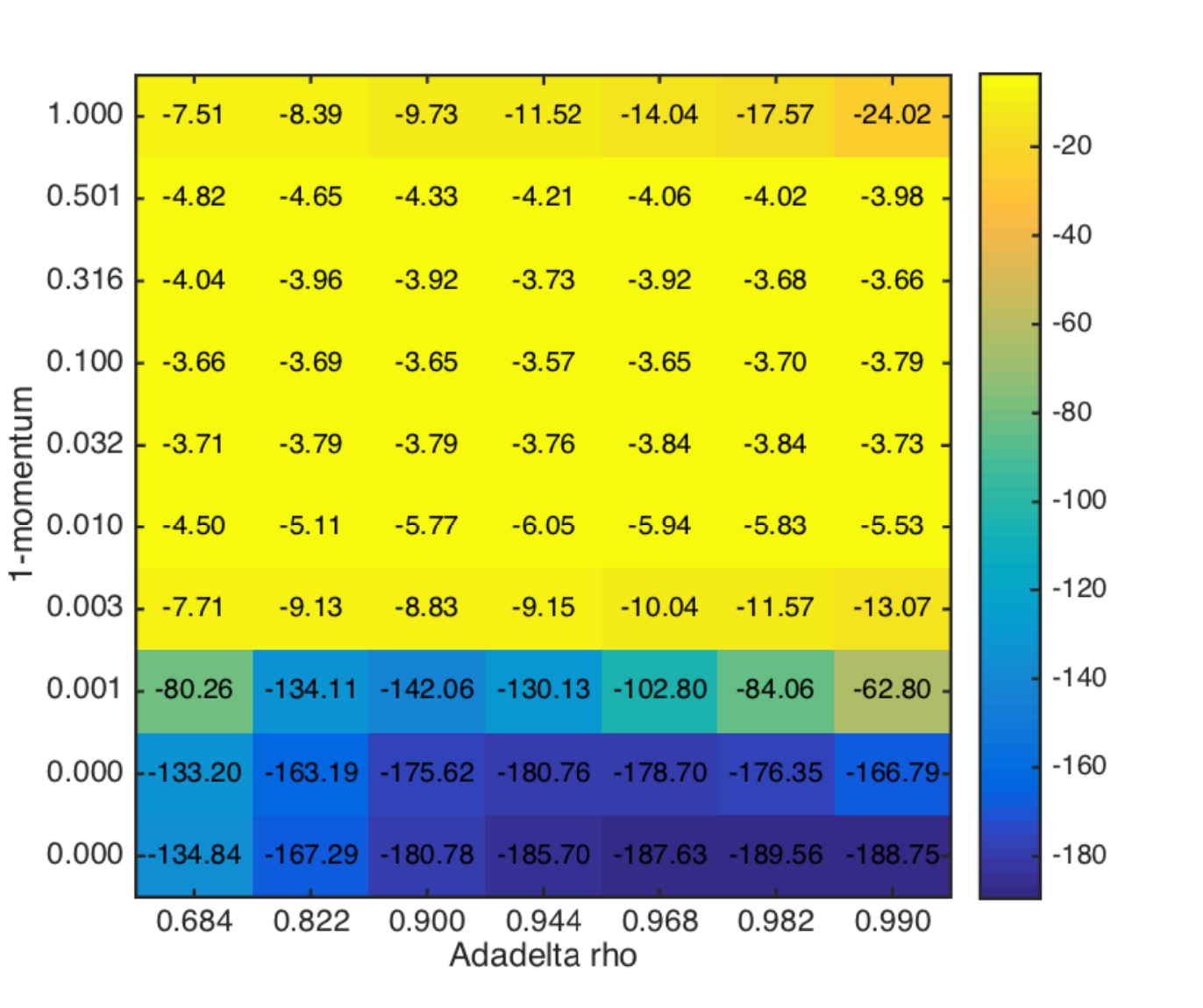}
\caption{Evidence lower bound $(/1000)$ for the edge partition model (EPM) on the NIPS dataset, achieved after 1000 iterations using Adadelta with different values of momentum and $\rho$.}
\label{tuneAdadeltaNetwork}
\end{center}
\end{minipage}\hfill
\begin{minipage}{0.45\textwidth}
\centering
\includegraphics[width=\textwidth]{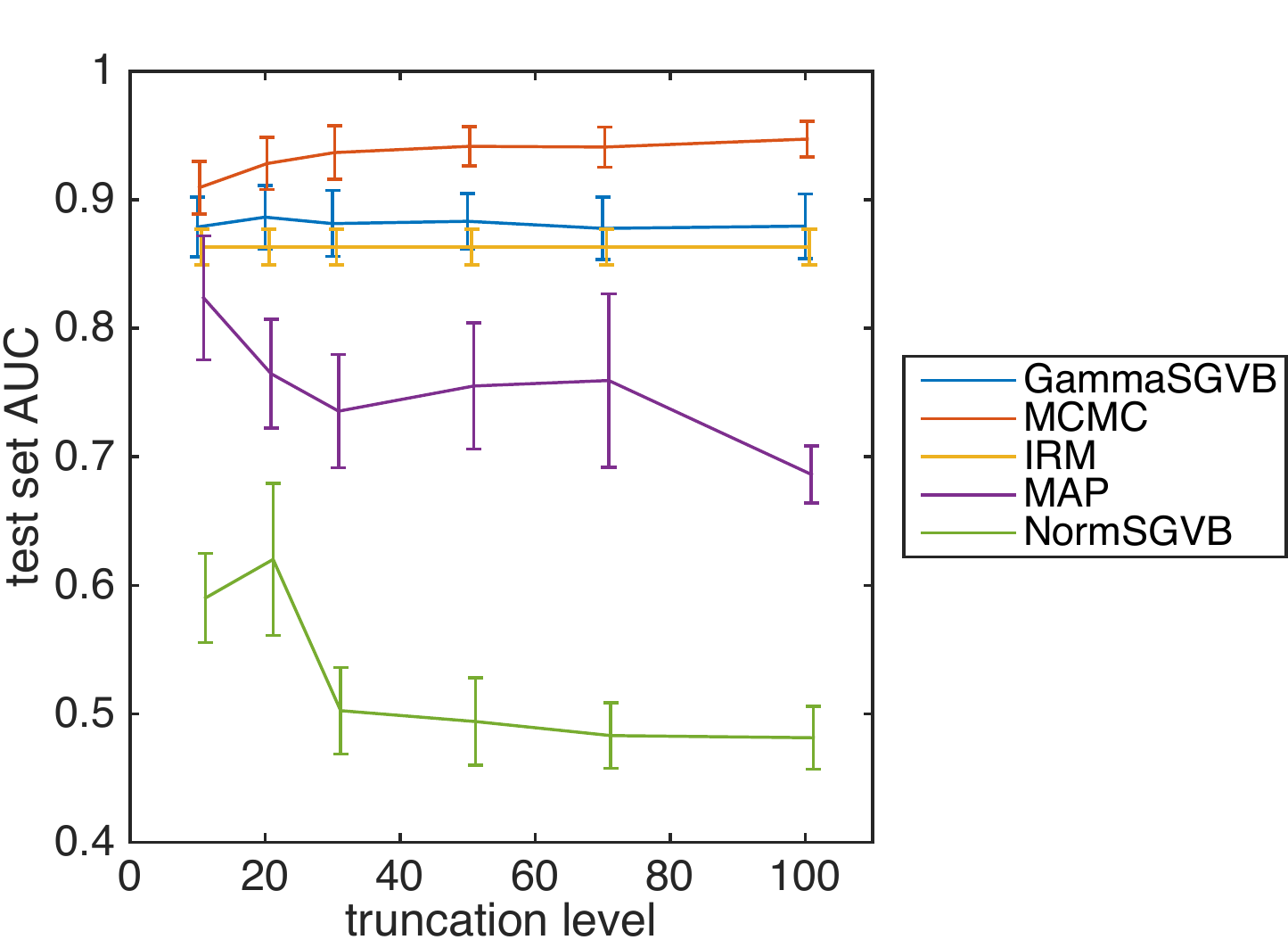}
\caption{Test set AUC for varying truncation level $K$ for the EPM on the NIPS dataset, across $10$ training/test splits.}
\label{fignips}
\end{minipage}
\end{figure}

Since Adadelta seemed the most promising of the ``automatic'' methods we sought to validate the claim that the optimisation performance is not particularly sensitive to the choice of $\rho$ and the momentum $\lambda$. Figure \ref{tuneAdadeltaNetwork} shows the ELBO achieved using Adadelta after 1000 iterations for varying $\rho$ and $\lambda$. For $1-\lambda$ in a range from $0.3$ to $0.03$ and $\rho$ across the full range tested ($0.684$ to $0.99$) the performance is very similar. Consider $1-\lambda=0.1$ i.e. $\lambda=0.9$: this is roughly equivalent to using information from the last $1/0.1=10$ samples to calculate the gradient, which seems intuitively reasonable given these are independent samples from $q$. In contrast $1-\lambda=1$ means that only the current gradient is used (i.e. no momentum) which we see degrades performance, implying the gradient estimates are then somewhat too noisy. 

In order to assess performance quantitatively we compare to the MCMC implementation from \citet{zhou2015infinite}, , and the infinite relational model, at link prediction on the NIPS $N=234$ dataset\footnote{\url{http://chechiklab.biu.ac.il/~gal/data.html}}. We attempted to compare to Stan, using Automatic Differentiation Variational Inference~\citep[AVDI,][]{kucukelbir2015automatic}, but the gradient evaluations were always \texttt{nan} at initialization. We use 10 training-test splits taking 20\% of pairs as test data, and report test set AUCs for varying truncation levels $K$ (Figure \ref{fignips}, note that the IRM is not truncated so its performance is equal at every $K$). While the carefully engineered MCMC algorithm consistently performs best (particularly for larger truncation levels), gammaSGVB still improves over the IRM. We include two alternative ``black box'' methods: MAP inference using gradient descent, and ``NormSGVB'' which is the equivalent algorithm to gammaSGVB but using a fully factorized normal distribution and using the reparameterization $r(\theta)=\log(1+\exp{(\theta}))$ to maintain non-negativity of the parameters, analogously to the approach used for SGVB in Stan. Both perform poorly, especially for larger truncation levels. In terms of runtime MCMC takes on average 30\% longer to run than gammaSGVB, but we emphasize that the MCMC implementation is tuned for this model, including for example model specific \texttt{mex} functions. By contrast, just 200 iterations of Hamiltonian Monte Carlo, which is ``black box'' in the same sense as our method of requiring only gradients, takes an order of magnitude longer than SGVB ($310$s vs $19$s for $K=10$) and still gives inferior performance (average test set AUC of $0.71$). 

\subsection{Gamma process factor analysis} 

\begin{figure}[htbp]
\centering
\begin{minipage}{0.66\textwidth}
\begin{center}
  \includegraphics[width=\textwidth]{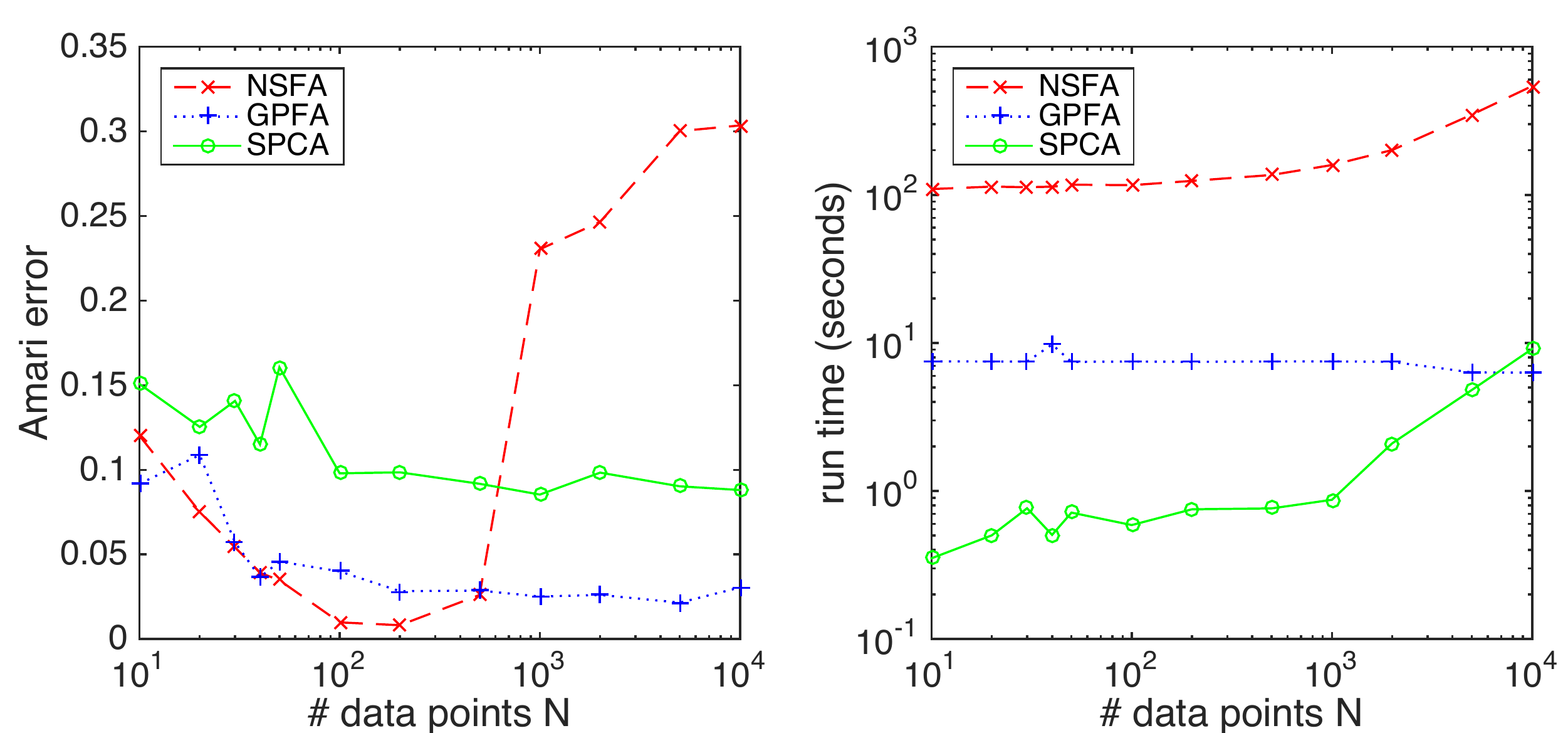}
\caption{Results on synthetic data for GPFA. \textbf{Left}: Amari error for reconstructing the factor loading matrix. \textbf{Right}: run time (1000 iterations/samples for GPFA/NSFA).}
\label{figSemiNMF}
\end{center}
\end{minipage}\hfill
\begin{minipage}{0.32\textwidth}
\begin{center}
  \includegraphics[width=\textwidth]{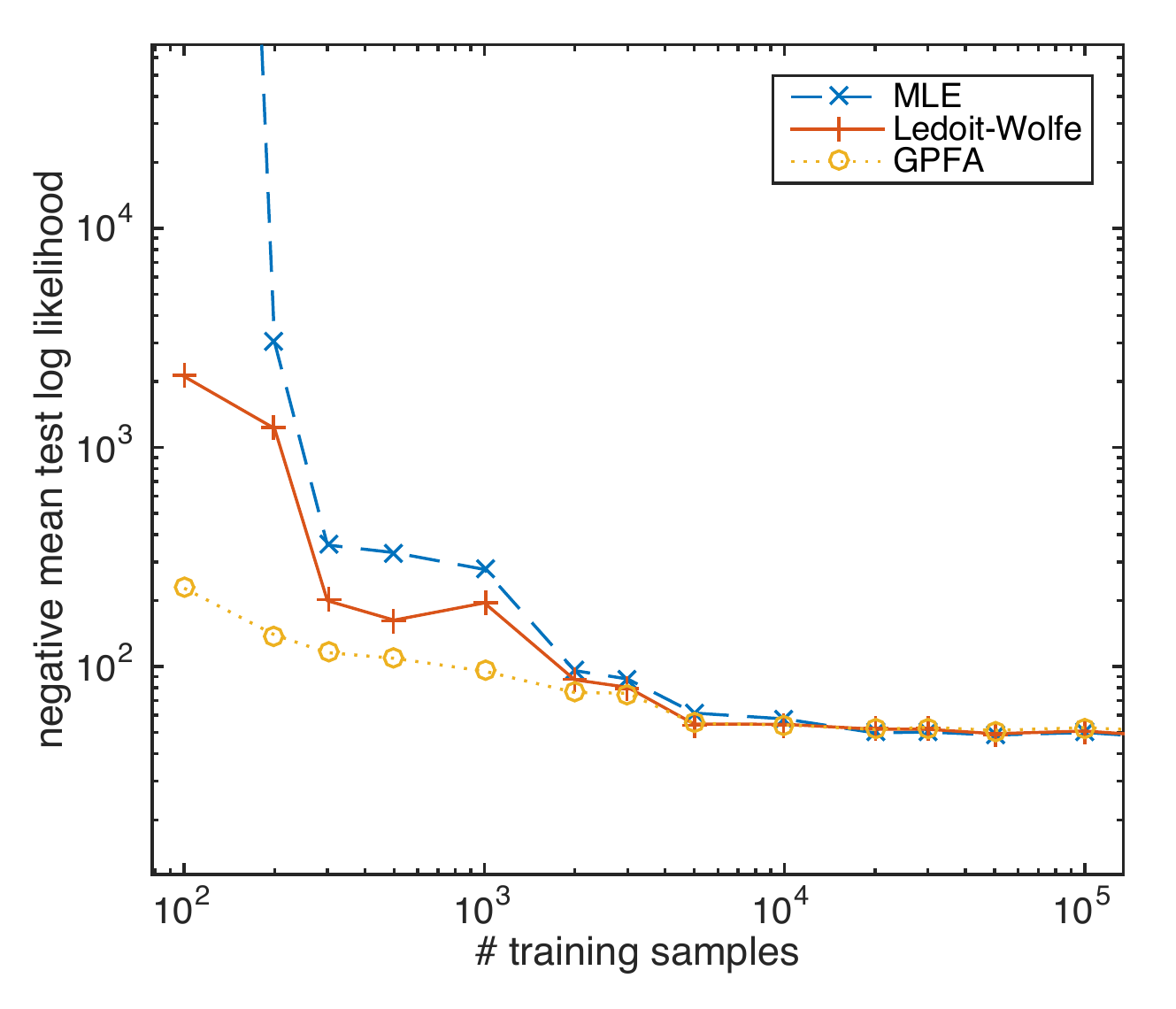}
\caption{Perplexity on CyTOF data with increasing training sample size. For $N=100$, the perplexity for the empirical covariance is $-10^{12}$. }
\label{figCytofPerf}
\end{center}
\end{minipage}
\end{figure}

We first test our implementation of GPFA using synthetic data. We fix the number of dimensions $D=50$, latent factors $K=10$ and vary the sample size $N$ from $10$ to $10^4$ to assess how the method copes with increasing sample size. The true factor loading matrix is sampled elementwise from the mixture $0.8 \delta_0 + 0.2 U[0,1]$, i.e. each element is non-zero with probability $0.2$, and those elements are uniform on $[0,1]$. The noise variance is $0.1$, and the true latent factors $\mathbf{x}_n \sim N(0,\mathbf{I})$. We compare to an MCMC implementation of the Indian Buffet Process based Nonparametric Sparse Factor Analysis \citep[NSFA,][]{knowles2011nonparametric} and the sparse PCA \citep[SPCA,][]{zou2006sparse} algorithm implemented in the SpaSM toolbox \citep{sjostrand2012spasm}. We allow SPCA to ``cheat'' by choosing the regularization parameter which minimizes the reconstruction error. To assess the recovery of the factor loadings $W$ we compute the Amari error \citep{amari1996new} which is invariant to permutation and scaling of the factor loadings. For small sample sizes $N \leq 200$, we see that NSFA typically slightly outperforms GPFA (Figure \ref{figSemiNMF}), presumably because the spike and slab prior better matches the true data generating mechanism. However, as $N$ increases the performance of NSFA actually degrades for the same number of MCMC iterations (1000), because convergence and mixing becomes problematic. In contrast the ability to integrate out the latent factors $X$ when using gammaSGVB means that the inference problem becomes easier rather than harder as the sample size increases. SPCA is consistently outperformed by GPFA, suggesting that the L1 regularization is not sufficient to reconstruct the factor loadings successfully, a finding that agrees with those in \citet{mohamed2011bayesian}. The computational cost of GPFA is also much lower than for NSFA because of the easily vectorized operations, and as noted in Section \ref{secGpfaModel}, GPFA's runtime has no dependence on $N$. The runtime of SPCA is approximately linear in $N$, so while it is considerably faster than GPFA for small $N$, by $N=10^4$ SPCA is actually slower. 

\begin{figure*}[htbp]
\begin{center}
  \includegraphics[width=\textwidth]{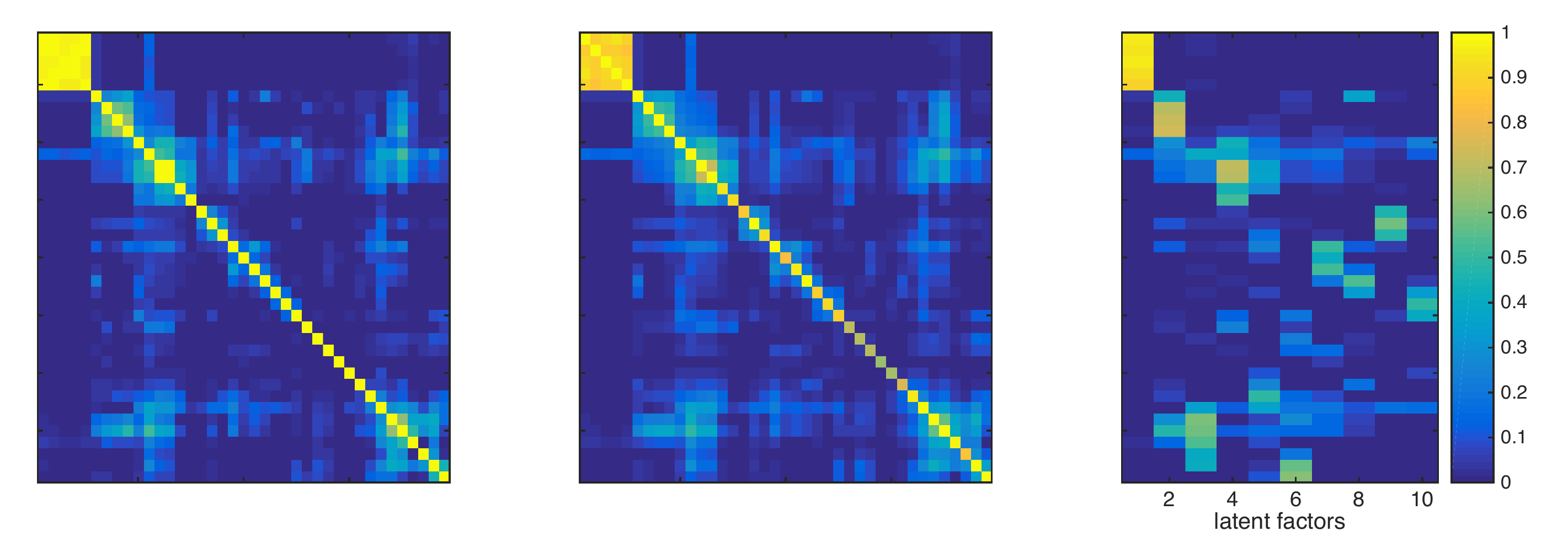}
\caption{Estimated covariance structure on CyTOF data. GPFA models the structure well, whilst regularizing the off diagonal components in particular. \textbf{Left}: empirical covariance. \textbf{Middle}: covariance estimated under GPFA. \textbf{Right}: top 10 latent factor loadings. These are easier to interpret than the usual PCA loadings because of the enforced non-negativity. }
\label{figCytof}
\end{center}
\end{figure*}

We apply GPFA to CyTOF~\citep{bendall2011single} data. CyTOF is a novel high through-put technology capable of measuring up to 40 protein abundance levels in thousands of individual
cells per second. Specific proteins are tagged using heavy metals which are measured using time-of-flight mass spectrometry. The sample we analyze consists of human immune cells, so representing the
heterogeneity between cells is relevant for understanding disease response. Our dataset has $N=5.3 \times 10^5$ cells and $D=40$ protein expression levels. We run GPFA for 3000 iterations using Adadelta($\rho=0.9, \epsilon=1 \times 10^{-4}$), $K=40$ and a prior $1/ \sigma^2 \sim G(.1,.1)$ on the noise variance. Runtime is around $10$ seconds on a quad-core 2.5GHz i7 MacBook Pro. To assess performance we split the dataset into a training and test set. Having fit the model on the training data, we calculate the perplexity (average negative log likelihood over test data points) of the remaining (test) data under the learnt model by drawing $S=100$ samples $W^{(s)}, \sigma^2_{(s)}  \sim q$, and obtaining the expected covariance matrix, 
\begin{align} \hat{Cov}(\mathbf{y}) = \frac1S \sum_{s=1}^S W^{(s)} W^{(s)} + \sigma^2_{(s)} I. \end{align}
We compare to two simple alternatives: using the maximum likelihood estimator (i.e. the empirical covariance), and Ledoit-Wolfe shrinkage \citep{ledoit2003improved}. We see that for fewer than around $N=2000$ training points GPFA outperforms the empirical covariance or Ledoit-Wolfe. In real datasets it is often the case that $N$ is not significantly larger than $D$, or even $N<D$ (usually referred as the ``large $p$, small $n$'' regime), so GPFA's strong performance for smaller sample sizes is valuable. Finally in Figure \ref{figCytof} we show the empirical and estimated covariances,  and the expected posterior factor loading matrix.

\section{Discussion} \label{sec:conclusion}

Variational inference has been considered a promising candidate for scaling Bayesian inference to real world datasets for some time. However, only with the advent of stochastic variational methods has this hope really started to become a reality. Alongside minibatch based SVI allowing improved scalability, by using Monte Carlo estimation SGVB can also allow a wider range of models to be easily handled than standard VBEM. The only model specific derivation we require is the gradient of the log joint, which is no more than is required for LBFGS or HMC. Indeed, automatic differentiation tools such as Theano \citep{Bastien-Theano-2012} could (and should!) be used to obtain these gradients. We have shown here that these ideas apply to sparse continuous latent variables represented using a gamma variational posterior, as well as to Gaussian variables. In addition we have shown that the ability to easily handle non-conjugate likelihoods can have advantages in terms of inference: in particular that collapsing models can improve performance. While this is well known for Latent Dirichlet Allocation \citep{blei2003latent}, leveraging this understanding has previously required careful model specific derivations (see e.g. \citet{teh2006collapsed}). An interesting potential line of future research would be to combine the ideas presented here with the variational autoencoder \citep{kingma2013auto} to allow scalable, nonlinear, sparse latent variable models, while additionally giving some ability to model posterior dependencies. 

\small
\bibliography{gammanmf}
\bibliographystyle{icml2015}

\end{document}